\documentclass[twoside]{article}

\usepackage{PRIMEarxiv}

\usepackage[utf8]{inputenc} 
\usepackage[T1]{fontenc}    
\usepackage{hyperref}       
\usepackage{url}            
\usepackage{booktabs}       
\usepackage{amsfonts}       
\usepackage{nicefrac}       
\usepackage{microtype}      
\usepackage{lipsum}
\usepackage{fancyhdr}       
\usepackage{graphicx}       

\usepackage{algorithm}

\usepackage{amssymb}
\usepackage{amsmath,amsfonts}
\usepackage{mathrsfs}

\usepackage{multirow}
\usepackage{makecell}
\usepackage{booktabs}
\usepackage{subfig}
\usepackage{algpseudocode}
\usepackage{siunitx}

\title{Performative Drift Resistant Classification using Generative Domain Adversarial Networks}

\author{
  Maciej Makowski, Brandon Gower-Winter, Georg Krempl \\
  Utrecht University \\
  8 Hiedelberglaan, Utrecht, 3584 CS, NL \\
  \texttt{m.w.makowski@students.uu.nl, b.gower-winter@uu.nl, g.m.krempl@uu.nl} \\
}

\begin{document}
\maketitle

\begin{abstract}
Performative Drift is a special type of Concept Drift that occurs when a model's predictions influence the future instances the model will encounter. In these settings, retraining is not always feasible. In this work, we instead focus on drift understanding as a method for creating drift-resistant classifiers. To achieve this, we introduce the Generative Domain Adversarial Network (GDAN) which combines both Domain and Generative Adversarial Networks. Using GDAN, domain-invariant representations of incoming data are created and a generative network is used to reverse the effects of performative drift.

Using semi-real and synthetic data generators, we empirically evaluate GDAN's ability to provide drift-resistant classification. Initial results are promising with GDAN limiting performance degradation over several timesteps. Additionally, GDAN's generative network can be used in tandem with other models to limit their performance degradation in the presence of performative drift. Lastly, we highlight the relationship between model retraining and the unpredictability of performative drift, providing deeper insights into the challenges faced when using traditional Concept Drift mitigation strategies in the performative setting.
\end{abstract}


\section{Introduction}

Machine Learning has transformed decision-making across various domains. However, once deployed, models may encounter unforeseen factors that degrade performance by requiring the model to make predictions on data that originated from a distribution different to its training data. This phenomena is more broadly known as Concept Drift \cite{GamaJoaoZliobaite2014} and is most commonly associated with environmental factors that are intrinsic to the setting the predictive model is deployed in. Recently, a new type of drift has been identified that does not arise due to these intrinsic factors, but rather due to the presence of the predictive model embedded within the setting it is making predictions in. Such scenarios are described as Performative \cite{PerdomoZrnicMendlerdunnerEtal2020}, and are identifiable by the causal relationship between the predictions a model makes and the drift the system experiences.

In such cases, performative drift can lead to dangerous feedback loops. For example, a company that struggles financially might face higher debt service costs (assigned by a predictive model) which, in turn, further engenders the company's financial struggles. To combat Concept Drift generally, methods like retraining with new data or online learning exist \cite{Hoi2018} but they are not always sustainable, especially in the performative setting where model updates are likely to introduce additional or different performative drift phenomena that must be perpetually reacted to. An alternative approach would be drift understanding \cite{Krempl2021beyond} whereby predictive models may be imbued with the ability to not only understand the drift they are experiencing, but adapt to it.

To achieve this goal, we propose a novel approach to handling model deterioration caused by performative drift. Leveraging domain adversarial neural networks (DANN) \cite{GaninUstinovaAjakan2016} and generative adversarial networks (GAN) \cite{CreswellAntoniaWhite2018, goodfellow2014}, our architecture, GDAN, extracts domain-invariant features, enabling more robust classification, regardless of whether the data originates from the original training distribution or one altered by model predictions.

Our main contributions are the following:
\begin{enumerate} 
\item The leveraging of drift understanding as a method for addressing performance degradation caused by performative drift. 
\item A novel architecture (GDAN) that is drift-resistant and capable of simulating distribution shifts caused by performative drift.
\item We demonstrate why performative drift is unique when compared to intrinsic drift, and illustrate why traditional methods for handling Concept Drift (such as incremental learning) may prove ineffective in performative settings.
\end{enumerate}

Lastly, we make the source code and Supplementary Materials for this paper available at: \url{https://tinyurl.com/erat3t47}

\section{Background and Related Work}
\subsection{Performativity}
If a setting is performative, a model's predictions affect the data that it must later predict on. Performative drift is a model-dependent form of Concept Drift and is distinct from traditional (intrinsic) concept drift which is model-independent. The term performativity was introduced described by Perdomo et al. \cite{PerdomoZrnicMendlerdunnerEtal2020} when they introduced the concept of performative risk, a loss function that describes the performance degradation a model will experience when evaluated on future data distributions it engenders. They also introduced \emph{repeated risk minimisation} (RRM), a procedure which repeatedly finds a model that minimizes the performative risk. Under strict assumptions, retraining in RRM can cause performative risk to converge, thus eliminating the need for model retraining.

Performativity has been discussed under various names in different domains. In the context of classification in data streams, the concept drift caused by this phenomenon has earlier been described as prediction-induced drift \cite{KremplBodnarHrubos2015IDA,Krempl2021beyond}. 
Performativity has also been observed in Recommender Systems, which have to manage the performative effect recommendations have on the content consumed by their users \cite{MansouryAbdollahpouriPechenizkiy2020, ChenDongWang2023, EpsteinHuangMegerdoomian2024}. Adversarial Learning is also inherently performative with adversaries updating their strategies in response to detectors which react and update their strategies in response to the adversaries \cite{BenJayaramWoodruff2022, KoryckiKrawczyk2024}.

\subsection{Domain Adversarial Neural Networks (DANNs)} 
Domain Adversarial Neural Networks (DANNs) \cite{GaninUstinovaAjakan2016} address domain adaptation by learning domain-invariant features that generalize across different data distributions. DANNs consist of three components: a feature extractor (F), a label predictor (LC), and a domain classifier (DC). The feature extractor aims to confuse the domain classifier by minimizing the following objective function: 

\begin{equation}
    E = \mathcal{L}_{LC} - \lambda \cdot \mathcal{L}_{DC}
\end{equation}

Training involves a min-max game where LC minimizes $\mathcal{L}_{LC}$, while DC tries to minimize $\mathcal{L}_{DC}$ and simultaneously maximize the overall objective.

\subsection{Generative Adversarial Networks} 
Generative Adversarial Networks (GANs) \cite{goodfellow2014} generate data resembling a given dataset by employing two neural networks: a generator (G) and a discriminator (D). The networks engage in a two-player min-max game, with the generator trying to produce fake samples $G(z)$ from noise $z \sim P(Z)$, while the discriminator tries to distinguish real data $x \sim P(X)$ from generated data. Training GANs can pose several challenges, such as training instability \cite{Luo2023Stabilizing}, undifferentiable outputs \cite{Kodali2017How} and mode collapse.
Standard GANs focus on distinguishing between real and fake data. Mirza et al. \cite{MirzaOsindero2014} enhanced this by conditioning the GAN on additional information, such as class labels, to generate more targeted outputs. Both the generator and discriminator receive extra inputs, enabling more controlled data generation. Building on this, Odena et al. \cite{OdenaOlah2017} introduced the Auxiliary Classifier GAN (AC-GAN), where the discriminator outputs two probability distributions: one for real vs. fake classification and another for class prediction. 
Isola et al. \cite{IsolaZhu2018} introduced a method for image-to-image translation using conditional GANs. The discriminator's input is extended by pairing real data $x$ with either target data $y$ or generator outputs $G(x)$, enabling the generator to learn how to produce target images from $x$. To improve results, the L1 distance metric is used between the translated image and the target image \cite{PathakDeepak2016}.

\section{GDAN - Generative Domain Adversarial Networks}

This section presents GDAN, a performative drift resistant architecture that maps new distributions back to their original pre-drift distributions. The system integrates Domain Adversarial Neural Networks and Generative Adversarial Networks, employing necessary modifications to combine these concepts.

\begin{figure}[t]
    \centering
    \includegraphics[width=0.7\textwidth]{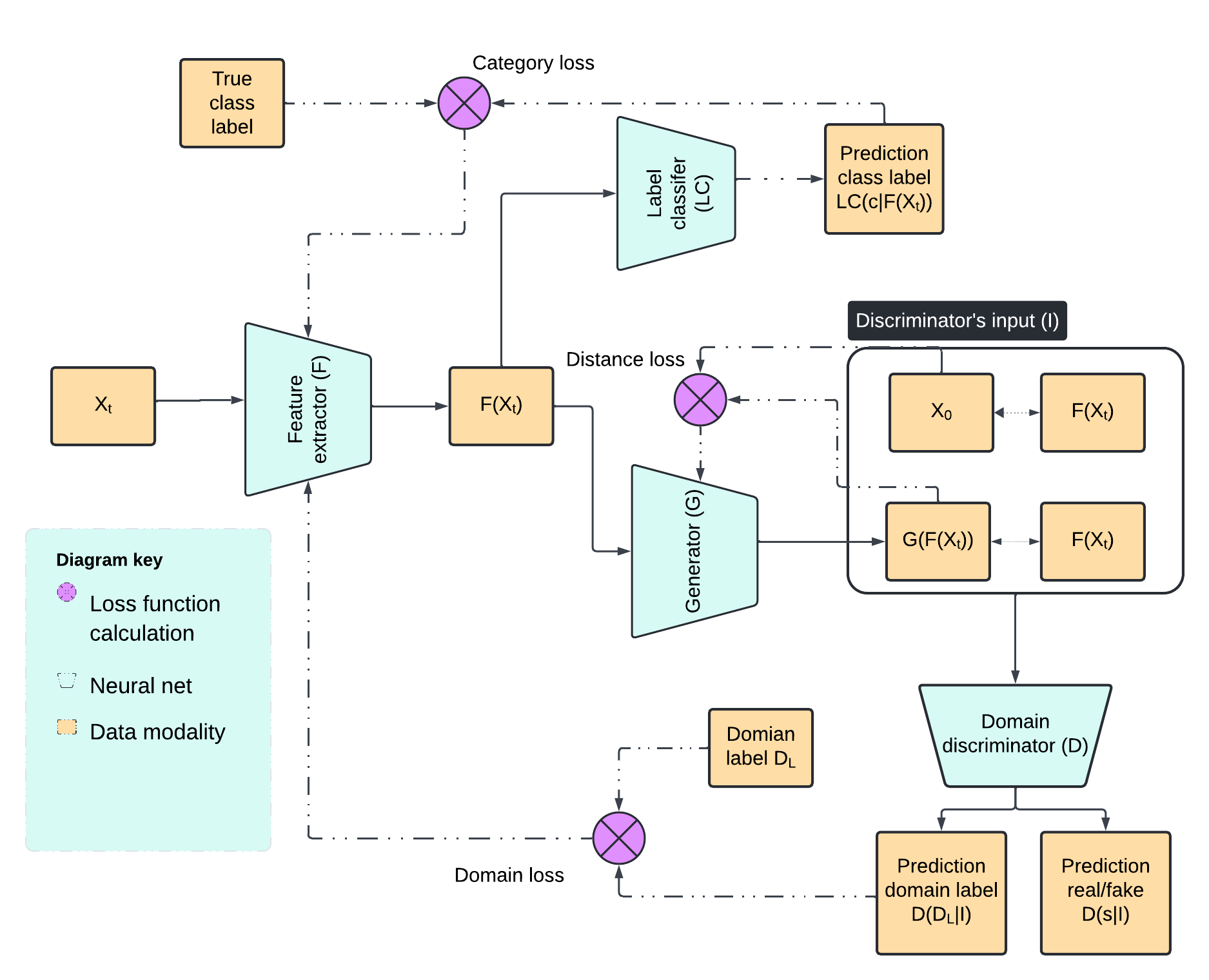}
    \caption{GDAN architecture, composed of a domain adversarial network and a generative adversarial network. Single-sided arrows illustrate an input/output relationship, while double-sided arrows describe concatenation of data.}
    \label{fig:GDANN_arch}
\end{figure}

Figure \ref{fig:GDANN_arch} illustrates the architecture of the model. Here, $X_t$ refers to a data point at iteration $t$. Each sample is represented as a tuple $(X_t, D_l, c)$. Where $X_t$ is the set of attributes, $D_l$ is the label of the distribution from which the point is sampled, and $c$ is the class label. GDAN comprises of the following components:

\begin{itemize}
    \item \textbf{Feature Extractor} ($F$): Extracts domain-invariant features from $X_t$, making the output $F(X_t)$ generic and domain-independent.
    \item \textbf{Label Classifier} ($LC$): Uses $F(X_t)$ to predict class $c$, ensuring robustness in classification despite distributional shifts caused by performative drift.
    \item \textbf{Generator} ($G$): Maps non-domain-specific representations $F(X_t)$ back to the original distribution. $G$ learns the nature of the drift from $X_0$ to $X_t$
    \item \textbf{Discriminator} ($D$): Input to the discriminator $I$ consists of real data points paired with their domain-invariant representations $(X_t, F(X_t))$ split equally with generated data points paired with the corresponding representations used to generate said data points $(G(F(X_t)), F(X_t))$. $D$ outputs two probability distributions: one indicating the origin of the data point ($D_l$) and the other assessing whether the point is from the real or generated distribution.
\end{itemize}

\subsection{Objective}
GDAN comprises of several components, all of which need to be considered when developing an objective function. First, consider the discriminator $D$. The loss of $D$ is two-fold: $L_s$ denotes the segment responsible for predicting whether the instance is sampled from $P(X)$, and $L_{d}$ denotes the log-likelihood of predicting the correct domain (i.e. which distribution was the point sampled from).

\begin{equation}
L_{s} = \mathbb{E}[log(P(S=1 |X_0, F(X_t))] + \mathbb{E} [log(P(S = 0 | G(F(X_t)), F(X_t))]
\end{equation}

\begin{equation}
L_{d} = \mathbb{E}[log(P(D_l=d_l|X_0, F(X_t))] + \mathbb{E}[log(P(D_l=d_l|G(F(X_t)), F(X_t))]
\end{equation}
\\
In Figure \ref{fig:GDANN_arch}, $I$ symbolizes the concatenated inputs supplied to $D$. The output of $D$ can then be written as $D(I) = (P(S|I), P(D_l|I))$. The goal of $D$ is to maximise $\mathcal{L}(G,D,F) = L_{s} + L_{d}$ while the goal of the $G$ is to maximise $L_{d} - L_{s}$. $G$ has two tasks: to fool $D$ and produce outputs which are similar to the target distribution. Therefore, $G$ needs to be penalised based on the distance (dissimilarity) between its output and target instances: $\mathcal{L}_{L1} (G)= \mathbb{E}_{X_0, X_t} [||X_0 - G(F(X_t)||_1] $. 

The task of selecting the optimal generator $G^*$, can be expressed as follows:
\begin{equation}
    G^* = arg \: \underset{G}{min} \:\underset{D}{max} \: ( \mathcal{L} (G,D,F) + \lambda \mathcal{L}_{L1}(G) )
    \label{eq:pix2pixdist_objectfun}
\end{equation}

Finally, let us consider the case  of the label classifier $LC$ and feature extractor $F$. Optimisation of $LC$ follows the classic supervised learning paradigm, whereby $LC$ aims to minimise its prediction loss $\mathcal{L}_{LC}(F,LC)$ by adjusting its parameters to accurately predict class labels based on features outputted by $F$. $F$ is trained using $\mathcal{L}_{LC}$ and $\mathcal{L}_{d}$. $F$ aims to maximise $LC$'s accuracy while confusing $D$, ensuring the extracted features are generic and domain-invariant.

\begin{equation}
    F^* = arg \: \underset{F}{min} \: \underset{D}{max} \: (\mathcal{L}_{LC} - \lambda \cdot \mathcal{L}_{P(D_l=d_l|X_0, F(X_t))} )
\end{equation}

To summarize the approach, during training, there are two adversarial two-player games. The first one involves the feature extractor $F$ and the part of the discriminator $D$ responsible for predicting the domain label, and the second one is the generator $G$ vs. the discriminator $D$. For further clarity, Pseudocode for training GDAN can be found in Appendix \ref{sect:PSEUDOCODE}.

\section{Experimental Setup}

This section describes our process for evaluating GDAN. By leveraging two data-generators developed by Perdomo et al. \cite{PerdomoZrnicMendlerdunnerEtal2020} and Izzo et al. \cite{IzzoZacharyZou2022}, we design a suite of experiments to compare the accuracy (performance) of GDAN against two benchmark models. We also describe how we evaluate GDAN's understanding of performative drift. Results from  a baseline experiment has been included in Appendix \ref{tab:BENCHMARK} and a full list of the network architectures and experimental parameters used in this work have been included in the Supplementary Materials. 

\subsection{Simulation Design and Evaluation Metrics} \label{sec:sim_design}
To the best of our knowledge, there exists no publicly available dataset with known performative drift. We are therefore required to use the semi-real Perdomo \cite{PerdomoZrnicMendlerdunnerEtal2020} and synthetic Izzo \cite{IzzoZacharyZou2022} data generators. Additionally, our experiments require a dataset with ground truth labels, where each data point is associated with a specific distribution. These distributions are sequentially generated by inducing performative drift on the preceding one. For performance evaluation, access to the parameters of the drift-inducing model is mandatory. 

A simulation run starts by generating data distribution $X_0$ using the chosen data generator. Then, a logistic regression model $M_0$ is trained on $X_0$ to establish performance benchmarks. $M_0$'s parameters are then given to the data generator, influencing the data distribution of future instances, simulating performative drift. Each iteration ($t$) represents a performative drift event. At each iteration, four models are evaluated: 

\begin{enumerate}
    \item \textbf{$M_0$}: The original logistic regression model trained on data $X_0$ and evaluated on data $X_i$.
    \item \textbf{$M_{ret}$}: A logistic regression model trained  and evaluated on data $X_i$.
    \item \textbf{$M_G$}: The original model $M_0$ evaluated on data generated by $G(F(X_t))$.
    \item \textbf{$M_{LC}$}: GDAN's label classifier.
\end{enumerate}

\begin{figure}[t]
    \centering
    \includegraphics[width=0.7\textwidth]{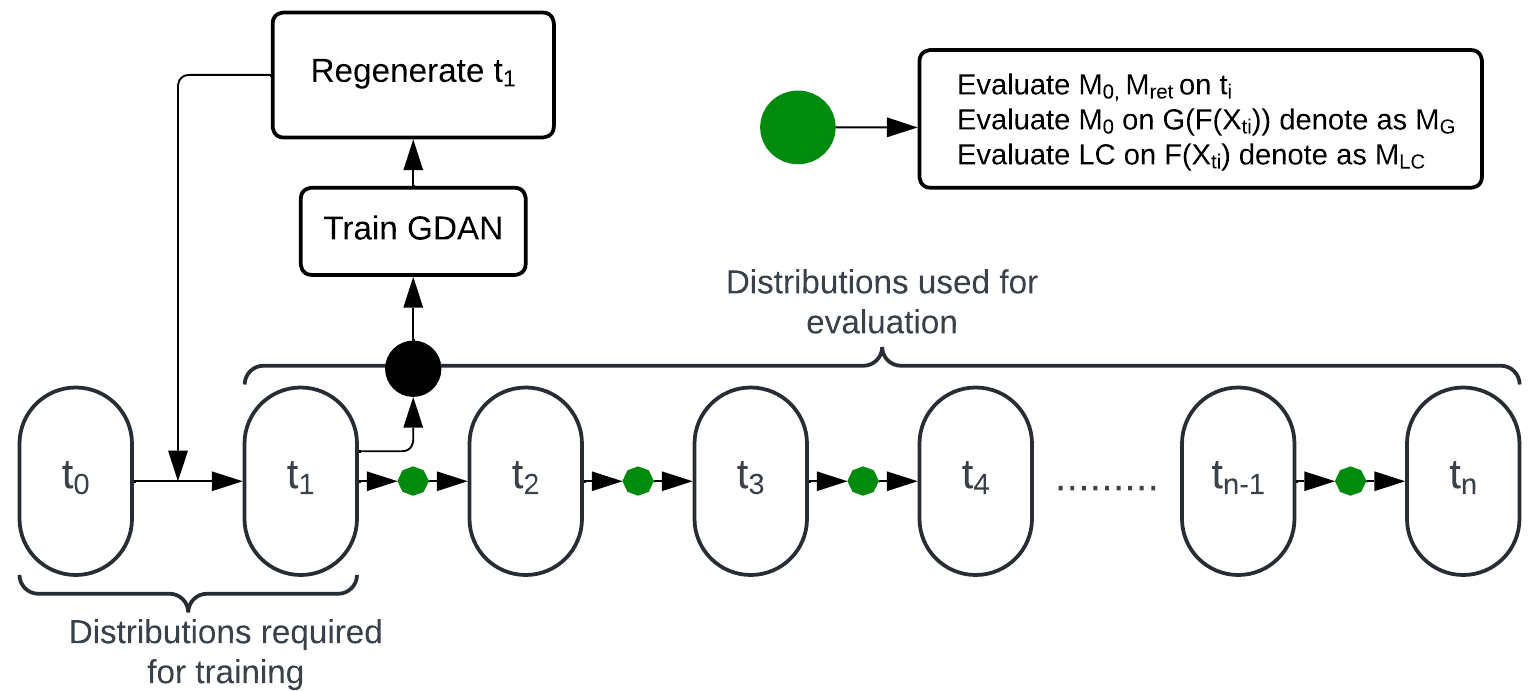}
    \caption{Visualization of the flow of the simulation. For training of the GDAN architecture, the only necessary distributions are annotated baseline $t=0$ and the first distribution when drift is detected $t=1$. Iterations $1 \ldots n$ are used for evaluation purposes. After a model is trained, $t=1$ is regenerated and the testing starts on it. In each iteration, four models are evaluated. The diagram depicts a division, which is similar to a test/train split in a classic ML setup.}
    \label{fig:test_train_split}
\end{figure}

At each iteration, prediction accuracy is recorded to ascertain the performance of each model. Additionally, The $L_1$ norm between $X_0$ and $X_i$ is compared to the $L_1$ norm between $X_0$ and $G(F(X_i))$. As a simulation run progresses, performative drift causes $X_i$ to look less like data distribution $X_0$, which should be reflected in the $L_1$ norm. If GDAN successfully learned a domain-invariant representation of the incoming data, the $L_1$ norm between $X_0$ and $G(F(X_i))$ should remain as close to $0.0$ as possible throughout the simulation run.

There is currently no framework for evaluating models in the performative setting. Thus, we treat the models $M_0$ and $M_{ret}$ as our baselines with $M_0$ representing a lower-bound given that it undergoes no retraining, while $M_{ret}$ acts like an upper-bound as it is continuously retrained after each drift event.

\subsection{Data generation}
\subsubsection{Perdomo Generator}
We implement a slightly modified version of the data generation process, described with the following equation.
\begin{equation}
    x_{t+1} = x_t - \epsilon B \theta
\end{equation}
$\theta$ symbolizes the feature coefficients of the logistic regression, $B$ indexes whether the feature is performative, i.e. it gets influenced by the drift. And $\epsilon$ informs about the strength of the performativity. $X_0$ is created by loading the data from \emph{GiveMeSomeCredit} dataset \cite{Kaggle2011}. For this experiment, we test two setups. One where $\theta$ is changed by retraining every iteration (Dynamic Drift), and second when $\theta$ is constant from $t=0$ (Monotonous Drift). 

\subsubsection{Izzo Generator}
For this generator we also slightly change the scheme of data generation, as we move it from $\mathbf{R}$ to $\mathbf{R^n}$. The main goal of this generator is to model a real-world spam detection situation, where only one class (the spammers) tries to modify their behaviour, to deceive the predictor.
\begin{equation}
    x|y=0, \theta \sim \mathcal{N}(\mu_0, \sigma_0^2) \qquad x|y=1, \theta \sim \mathcal{N}(f(\theta), \sigma_1^2)
\end{equation}
The function of moving the average of the distribution can be described with the following equation, where $n$ is the index of the feature. 
\begin{equation}
    \mu_{i,n} = f(\theta_n) = \mu_1 - \epsilon \cdot \theta_n \cdot B_n
    \label{eq:movement_description_extended}
\end{equation}

\section{Results and Discussion}

Table \ref{tab:results} presents the results of the experiments with the Perdomo data generator. The accuracies of $M_{LC}$ and $M_G$ are competitive in earlier iterations. Later they start diverging from the baseline, which is more rapid in the case where the drift is dynamic. Given that GDAN is trained on only two distributions of data, it only learns to replicate one direction of drift. Performance is better when the drift is monotonous as the direction of drift is constant. In the dynamic case, the direction of the drift is not constant which makes GDAN's task more complex. This is reflected in the larger standard deviations observed when the drift is dynamic. As the newly created distributions become increasingly different from the original distributions used for training, GDAN's performance degrades. However, the accuracy of $M_{LC}$ in the first five (monotonous), and first three (dynamic) iterations show that it is possible to train a classifier whose performance is superior to a retrained predictor. Additionally, the performance of $M_G$ indicates that GDAN can create domain-invariant representations which are understandable by models only trained on data from $t=0$ (e.g. $M_0$).

\renewcommand{\arraystretch}{1}

\begin{table}[t] 
\centering
\caption{Summary of experiments with Perdomo generator. Each experiment has been performed 10 times, and each time we sample 10000 points from the data generator and evolve them by inducing drift. The values presented in the table are means and standard deviations.}
\resizebox{\textwidth}{!}{
\begin{tabular}{@{}p{1.8cm}|p{3cm}|p{1cm}p{1cm}p{1cm}p{1cm}p{1cm}p{1cm}p{1cm}p{1cm}p{1cm}p{1cm}@{}}
\toprule
\textbf{Method} & \textbf{Metric} 
& \textbf{1} & \textbf{2} & \textbf{3} & \textbf{4} & \textbf{5} & \textbf{6} & \textbf{7} & \textbf{8} & \textbf{9} & \textbf{10} \\ \midrule

\multirow{6}{*}{\parbox{2.5cm}{\raggedright \textbf{Monot- \\onous \\ Drift}}} 
    & Acc $M_0$ & 
        \makecell{\textbf{71.53} \\ $\pm0.285$} & \makecell{\textbf{71.04} \\ $\pm0.311$} & \makecell{\textbf{70.05} \\ $\pm0.291$} & \makecell{\textbf{68.92} \\ $\pm0.274$} & \makecell{\textbf{67.92} \\ $\pm0.285$} & \makecell{\textbf{66.91} \\ $\pm0.303$} & \makecell{\textbf{66.04} \\ $\pm0.329$} & \makecell{\textbf{65.13} \\ $\pm0.334$} & \makecell{\textbf{64.14} \\ $\pm0.327$} & \makecell{\textbf{63.22} \\ $\pm0.303$} \\
        
    & Acc $M_G$ & 
        \makecell{\textbf{72.71} \\ $\pm0.268$} & \makecell{\textbf{72.26} \\ $\pm0.206$} & \makecell{\textbf{71.96} \\ $\pm0.178$} & \makecell{\textbf{71.88} \\ $\pm0.207$} & \makecell{\textbf{71.80} \\ $\pm0.180$} & \makecell{\textbf{71.56} \\ $\pm0.167$} & \makecell{\textbf{70.88} \\ $\pm0.142$} & \makecell{\textbf{70.06} \\ $\pm0.236$} & \makecell{\textbf{68.94} \\ $\pm0.249$} & \makecell{\textbf{67.53} \\ $\pm0.341$} \\
        
    & Acc $M_{LC}$ & 
        \makecell{\textbf{76.02} \\ $\pm0.302$} & \makecell{\textbf{75.92} \\ $\pm0.279$} & \makecell{\textbf{74.91} \\ $\pm0.198$} & \makecell{\textbf{73.59} \\ $\pm0.203$} & \makecell{\textbf{71.91} \\ $\pm0.177$} & \makecell{\textbf{69.55} \\ $\pm0.209$} & \makecell{\textbf{66.63} \\ $\pm0.337$} & \makecell{\textbf{63.38} \\ $\pm0.273$} & \makecell{\textbf{59.94} \\ $\pm0.269$} & \makecell{\textbf{56.78} \\ $\pm0.286$} \\
        
    & Acc $M_{ret}$ & 
        \makecell{\textbf{71.38} \\ $\pm1.480$} & \makecell{\textbf{71.38} \\ $\pm1.485$} & \makecell{\textbf{71.38} \\ $\pm1.481$} & \makecell{\textbf{71.38} \\ $\pm1.487$} & \makecell{\textbf{71.38} \\ $\pm1.482$} & \makecell{\textbf{71.38} \\ $\pm1.484$} & \makecell{\textbf{71.38} \\ $\pm1.480$} & \makecell{\textbf{71.38} \\ $\pm1.482$} & \makecell{\textbf{71.38} \\ $\pm1.489$} & \makecell{\textbf{71.38} \\ $\pm1.485$} \\
        
    & $L_1(X_0, X_i)$ & 
        \makecell{\textbf{0.107} \\ $\pm0.000$} & \makecell{\textbf{0.213} \\ $\pm0.000$} & \makecell{\textbf{0.32} \\ $\pm0.000$} & \makecell{\textbf{0.427} \\ $\pm0.000$} & \makecell{\textbf{0.533} \\ $\pm0.000$} & \makecell{\textbf{0.64} \\ $\pm0.000$} & \makecell{\textbf{0.747} \\ $\pm0.000$} & \makecell{\textbf{0.853} \\ $\pm0.000$} & \makecell{\textbf{0.96} \\ $\pm0.000$} & \makecell{\textbf{1.067} \\ $\pm0.000$} \\
        
    & $L_1(X_0, G(F(X_i))$ & 
        \makecell{\textbf{0.112} \\ $\pm0.001$} & \makecell{\textbf{0.119} \\ $\pm0.001$} & \makecell{\textbf{0.136} \\ $\pm0.001$} & \makecell{\textbf{0.158} \\ $\pm0.001$} & \makecell{\textbf{0.182} \\ $\pm0.001$} & \makecell{\textbf{0.204} \\ $\pm0.001$} & \makecell{\textbf{0.226} \\ $\pm0.001$} & \makecell{\textbf{0.247} \\ $\pm0.001$} & \makecell{\textbf{0.266} \\ $\pm0.001$} & \makecell{\textbf{0.285} \\ $\pm0.001$}\\
        \midrule
\multirow{6}{*}{\parbox{2.5cm}{\raggedright \textbf{Dynamic \\ Drift}}} 
    & Acc $M_0$ & 
        \makecell{\textbf{71.79} \\ $\pm0.387$} & \makecell{\textbf{71.2} \\ $\pm0.431$} & \makecell{\textbf{70.17} \\ $\pm0.655$} & \makecell{\textbf{69.07} \\ $\pm0.774$} & \makecell{\textbf{68.52} \\ $\pm0.896$} & \makecell{\textbf{66.96} \\ $\pm0.953$} & \makecell{\textbf{66.04} \\ $\pm1.043$} & \makecell{\textbf{65.12} \\ $\pm1.252$} & \makecell{\textbf{64.19} \\ $\pm1.284$} & \makecell{\textbf{63.38} \\ $\pm1.255$} \\
        
    & Acc $M_G$ & 
        \makecell{\textbf{72.8} \\ $\pm0.261$} & \makecell{\textbf{72.29} \\ $\pm0.420$} & \makecell{\textbf{71.89} \\ $\pm0.456$} & \makecell{\textbf{71.5} \\ $\pm0.474$} & \makecell{\textbf{70.71} \\ $\pm1.143$} & \makecell{\textbf{69.98} \\ $\pm1.874$} & \makecell{\textbf{69.3} \\ $\pm2.277$} & \makecell{\textbf{68.3} \\ $\pm2.672$} & \makecell{\textbf{67.05} \\ $\pm2.990$} & \makecell{\textbf{65.83} \\ $\pm3.292$} \\
        
    & Acc $M_{LC}$ & 
        \makecell{\textbf{75.99} \\ $\pm0.264$} & \makecell{\textbf{75.35} \\ $\pm0.832$} & \makecell{\textbf{72.32} \\ $\pm3.793$} & \makecell{\textbf{68.34} \\ $\pm6.717$} & \makecell{\textbf{64.3} \\ $\pm8.599$} & \makecell{\textbf{60.97} \\ $\pm9.147$} & \makecell{\textbf{58.21} \\ $\pm9.122$} & \makecell{\textbf{55.59} \\ $\pm8.515$} & \makecell{\textbf{53.35} \\ $\pm7.729$} & \makecell{\textbf{51.62} \\ $\pm6.976$} \\
        
    & Acc $M_{ret}$ & 
        \makecell{\textbf{72.29} \\ $\pm0.945$} & \makecell{\textbf{72.29} \\ $\pm0.945$} & \makecell{\textbf{72.29} \\ $\pm0.947$} & \makecell{\textbf{72.29} \\ $\pm0.947$} & \makecell{\textbf{72.29} \\ $\pm0.949$} & \makecell{\textbf{72.29} \\ $\pm0.947$} & \makecell{\textbf{72.29} \\ $\pm0.945$} & \makecell{\textbf{72.29} \\ $\pm0.938$} & \makecell{\textbf{72.29} \\ $\pm0.952$} & \makecell{\textbf{72.29} \\ $\pm0.944$} \\
        
    & $L_1(X_0, X_i)$ & 
        \makecell{\textbf{0.107} \\ $\pm0.000$} & \makecell{\textbf{0.232} \\ $\pm0.022$} & \makecell{\textbf{0.36} \\ $\pm0.041$} & \makecell{\textbf{0.488} \\ $\pm0.061$} & \makecell{\textbf{0.616} \\ $\pm0.081$} & \makecell{\textbf{0.744} \\ $\pm0.010$} & \makecell{\textbf{0.872} \\ $\pm0.012$} & \makecell{\textbf{1.0} \\ $\pm0.141$} & \makecell{\textbf{1.128} \\ $\pm0.161$} & \makecell{\textbf{1.255} \\ $\pm0.180$} \\
        
    & $L_1(X_0,G(F(X_i))$ & 
        \makecell{\textbf{0.112} \\ $\pm0.001$} & \makecell{\textbf{0.127} \\ $\pm0.008$} & \makecell{\textbf{0.155} \\ $\pm0.015$} & \makecell{\textbf{0.186} \\ $\pm0.021$} & \makecell{\textbf{0.216} \\ $\pm0.025$} & \makecell{\textbf{0.245} \\ $\pm0.031$} & \makecell{\textbf{0.275} \\ $\pm0.037$} & \makecell{\textbf{0.304} \\ $\pm0.046$} & \makecell{\textbf{0.335} \\ $\pm0.055$} & \makecell{\textbf{0.367} \\ $\pm0.065$} \\
        \bottomrule
\end{tabular}}
\label{tab:results}
\end{table}

Considering the distance ($L_1$) metrics, it is noticeable that the data generator with dynamic performative drift produces data with greater dissimilarity from iteration $t=0$ to $t=i$ when compared to the data generator with monotonous performative drift. Additionally, GDAN's reconstructed distributions $G(F(X_{i}))$ are always closer to the original distribution $X_0$, indicating an ability to reverse the distribution shifts caused by the performative drift. 

These results demonstrate one of the key characteristics of performative drift. Whenever a model is retrained, the performative drift's behaviour changes too. Retraining is a commonplace solution for handling intrinsic drift. Our results suggest that retraining would work in the performative setting if the deployed model has unfettered access to the labels of the instances it predicts on. If that is not case, our results indicate that less frequent retraining with a greater focus on drift understanding and drift reversal (using GDAN for example) may reduce the rate of performance degradation (See $M_g$ compared to $M_0$ in Table \ref{tab:results} for example). This is because fewer retraining steps would provide more monotonous performative drift which our results clearly show is easier to manage.

\renewcommand{\arraystretch}{1}

\begin{table}[t]
    \centering
    \caption{Summary of experiments with Izzo data generator (20k datapoints sampled, averaged over 20 runs). $M_{LC}$ and $M_{ret}$ are excluded, as both achieve perfect accuracy. All values oscillate with the retraining of the logistic regressor (every 3 iterations), demonstrating the cyclic nature of the performative drift.}
    \resizebox{\textwidth}{!}{
    \begin{tabular}{@{}p{2.5cm}|p{1cm}p{1cm}p{1cm}p{1cm}p{1cm}p{1cm}p{1cm}p{1cm}p{1cm}p{1cm}p{1cm}@{}}
    \toprule
        Metric/iter & 1 & 2 & 3 & 4 & 5 & 6 & 7 & 8 & 9 & 10 \\
        \midrule
        Acc $M_0$ & 
            \makecell{\textbf{50.3} \\  $\pm0.535$} & \makecell{\textbf{50.21} \\ $\pm0.295$} & \makecell{\textbf{50.02} \\ $\pm0.349$} & \makecell{\textbf{50.21} \\ $\pm0.449$} & \makecell{\textbf{60.03} \\ $\pm19.98$} & \makecell{\textbf{60.11} \\ $\pm19.95$} & \makecell{\textbf{59.88} \\ $\pm20.06$} & \makecell{\textbf{49.99} \\ $\pm0.372$} & \makecell{\textbf{50.21} \\ $\pm0.436$} & \makecell{\textbf{49.94} \\ $\pm0.431$} \\
        
        Acc $M_G$ & 
            \makecell{\textbf{78.59} \\ $\pm18.36$} & \makecell{\textbf{78.52} \\ $\pm18.34$} & \makecell{\textbf{78.63} \\ $\pm18.28$} & \makecell{\textbf{78.63} \\ $\pm18.21$} & \makecell{\textbf{94.33} \\ $\pm12.54$} & \makecell{\textbf{94.47} \\ $\pm12.26$} & \makecell{\textbf{94.24} \\ $\pm12.74$} & \makecell{\textbf{77.12} \\ $\pm18.72$} & \makecell{\textbf{77.07} \\ $\pm18.78$} & \makecell{\textbf{77.17} \\ $\pm18.68$} \\
            
        $L_1(X_0,X_i)$ & 
            \makecell{\textbf{3.78} \\ $\pm0.065$} & \makecell{\textbf{3.78} \\ $\pm0.068$} & \makecell{\textbf{3.78} \\ $\pm0.065$} & \makecell{\textbf{3.78} \\ $\pm0.066$} & \makecell{\textbf{3.67} \\ $\pm0.143$} & \makecell{\textbf{3.67} \\ $\pm0.138$} & \makecell{\textbf{3.66} \\ $\pm0.142$} & \makecell{\textbf{3.82} \\ $\pm0.096$} & \makecell{\textbf{3.82} \\ $\pm0.090$} & \makecell{\textbf{3.82} \\ $\pm0.091$} \\
        
        $L_1(X_0, G(F(X_i))$ & 
        \makecell{\textbf{3.39} \\ $\pm0.024$} & \makecell{\textbf{3.39} \\ $\pm0.017$} & \makecell{\textbf{3.39} \\ $\pm0.014$} & \makecell{\textbf{3.39} \\ $\pm0.012$} & \makecell{\textbf{3.41} \\ $\pm0.028$} & \makecell{\textbf{3.39} \\ $\pm0.026$} & \makecell{\textbf{3.39} \\ $\pm0.014$} & \makecell{\textbf{3.4} \\ $\pm0.042$} & \makecell{\textbf{3.4} \\ $\pm0.031$} & \makecell{\textbf{3.39} \\ $\pm0.030$} \\
        
        \midrule
        
        Metric/iter & 11 & 12 & 13 & 14 & 15 & 16 & 17 & 18 & 19 & 20 \\ 
        \midrule
        Acc $M_0$ &
            \makecell{\textbf{65.1} \\ $\pm22.76$} & \makecell{\textbf{64.88} \\ $\pm22.91$} & \makecell{\textbf{64.73} \\ $\pm23.01$} & \makecell{\textbf{49.83} \\ $\pm0.700$} & \makecell{\textbf{50.15} \\ $\pm0.650$} & \makecell{\textbf{50.14} \\ $\pm0.287$} & \makecell{\textbf{63.66} \\ $\pm21.06$} & \makecell{\textbf{63.74} \\ $\pm21.10$} & \makecell{\textbf{63.57} \\ $\pm21.13$} & \makecell{\textbf{50.13} \\ $\pm0.289$} \\
        
        Acc $M_G$ &
            \makecell{\textbf{96.42} \\ $\pm10.73$} & \makecell{\textbf{96.32} \\ $\pm11.02$} & \makecell{\textbf{96.35} \\ $\pm10.94$} & \makecell{\textbf{76.07} \\ $\pm19.05$} & \makecell{\textbf{76.57} \\ $\pm18.68$} & \makecell{\textbf{76.44} \\ $\pm18.78$} & \makecell{\textbf{99.56} \\ $\pm1.070$} & \makecell{\textbf{99.53} \\ $\pm1.149$} & \makecell{\textbf{99.56} \\ $\pm1.033$} & \makecell{\textbf{74.54} \\ $\pm17.57$} \\
            
        $L_1(X_0,X_i)$ & 
            \makecell{\textbf{3.67} \\ $\pm0.133$} & \makecell{\textbf{3.67} \\ $\pm0.132$} & \makecell{\textbf{3.68} \\ $\pm0.134$} & \makecell{\textbf{3.79} \\ $\pm0.030$} & \makecell{\textbf{3.79} \\ $\pm0.033$} & \makecell{\textbf{3.79} \\ $\pm0.035$} & \makecell{\textbf{3.68} \\ $\pm0.139$} & \makecell{\textbf{3.69} \\ $\pm0.137$} & \makecell{\textbf{3.69} \\ $\pm0.130$} & \makecell{\textbf{3.79} \\ $\pm0.037$} \\
            
        $L_1(X_0, G(F(X_i))$ & 
            \makecell{\textbf{3.38} \\ $\pm0.021$} & \makecell{\textbf{3.38} \\ $\pm0.021$} & \makecell{\textbf{3.39} \\ $\pm0.019$} & \makecell{\textbf{3.39} \\ $\pm0.019$} & \makecell{\textbf{3.39} \\ $\pm0.019$} & \makecell{\textbf{3.38} \\ $\pm0.026$} & \makecell{\textbf{3.38} \\ $\pm0.019$} & \makecell{\textbf{3.38} \\ $\pm0.036$} & \makecell{\textbf{3.39} \\ $\pm0.024$} & \makecell{\textbf{3.39} \\ $\pm0.015$}\\
    \bottomrule
    \end{tabular}
    }
    \label{tab:Izzo_results}
\end{table}

Table \ref{tab:Izzo_results} presents the results of the Izzo experiments. Interestingly, the nature of the Izzo generator's drift is cyclic. This can be seen by the oscillation of $M_0$ and $M_G$'s accuracies every three iterations. Additionally, the original logistic regression achieves greater accuracy on the mapped representations generated by GDAN ($M_G$). This indicates that despite GDAN's inability to completely minimize the dissimilarity of the drifted data points, GDAN is still capable of providing drift resistant classification. This is further supported by GDAN's label classifier which achieved perfect accuracy across all simulation runs.

\begin{figure}[t]

\centering
\subfloat[Iteration 0]{\label{fig:KDE_iter_1}
\centering
\includegraphics[width=0.49\linewidth]{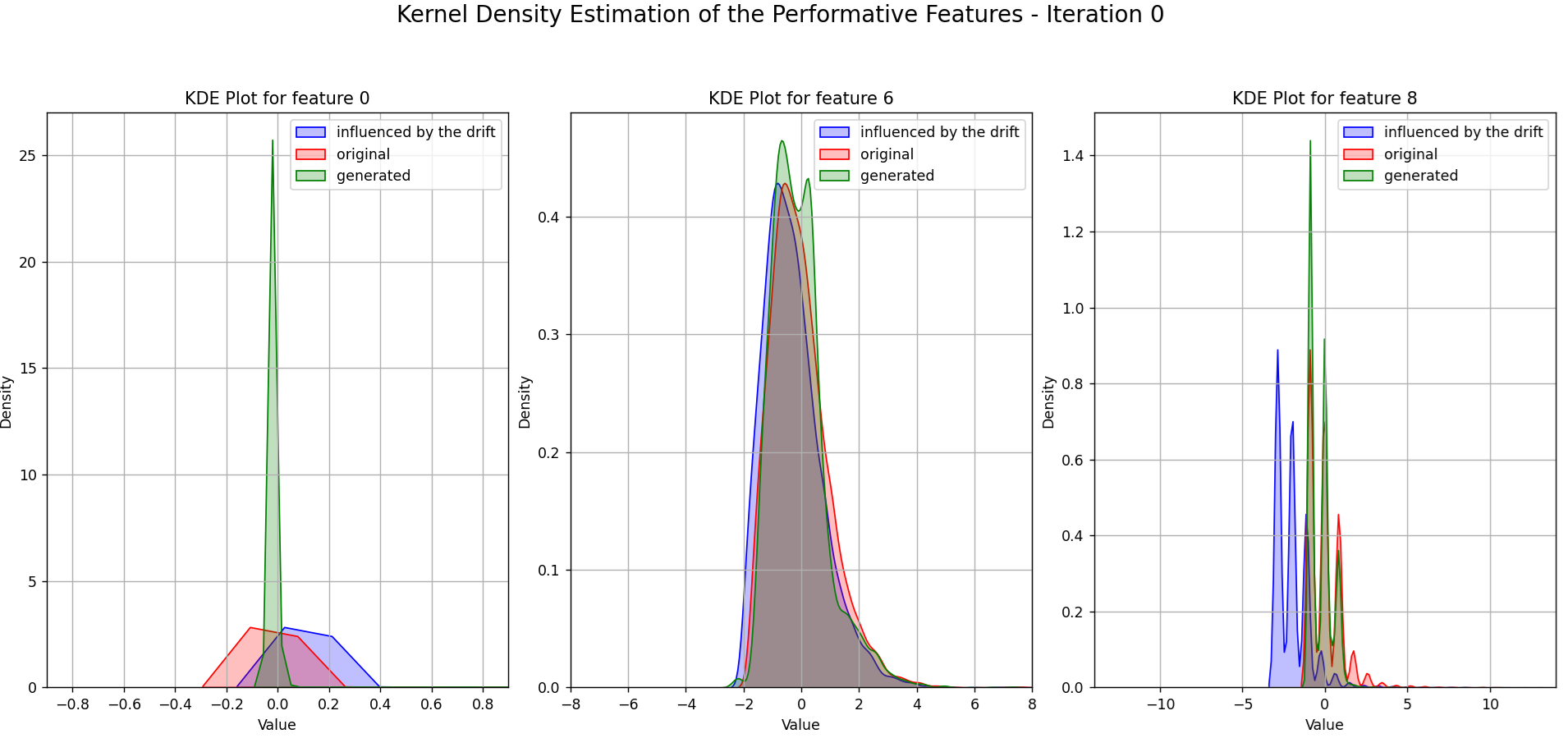}
}
\subfloat[Iteration 9]{\label{fig:KDE_iter_9}
\centering

\includegraphics[width=0.49\linewidth]{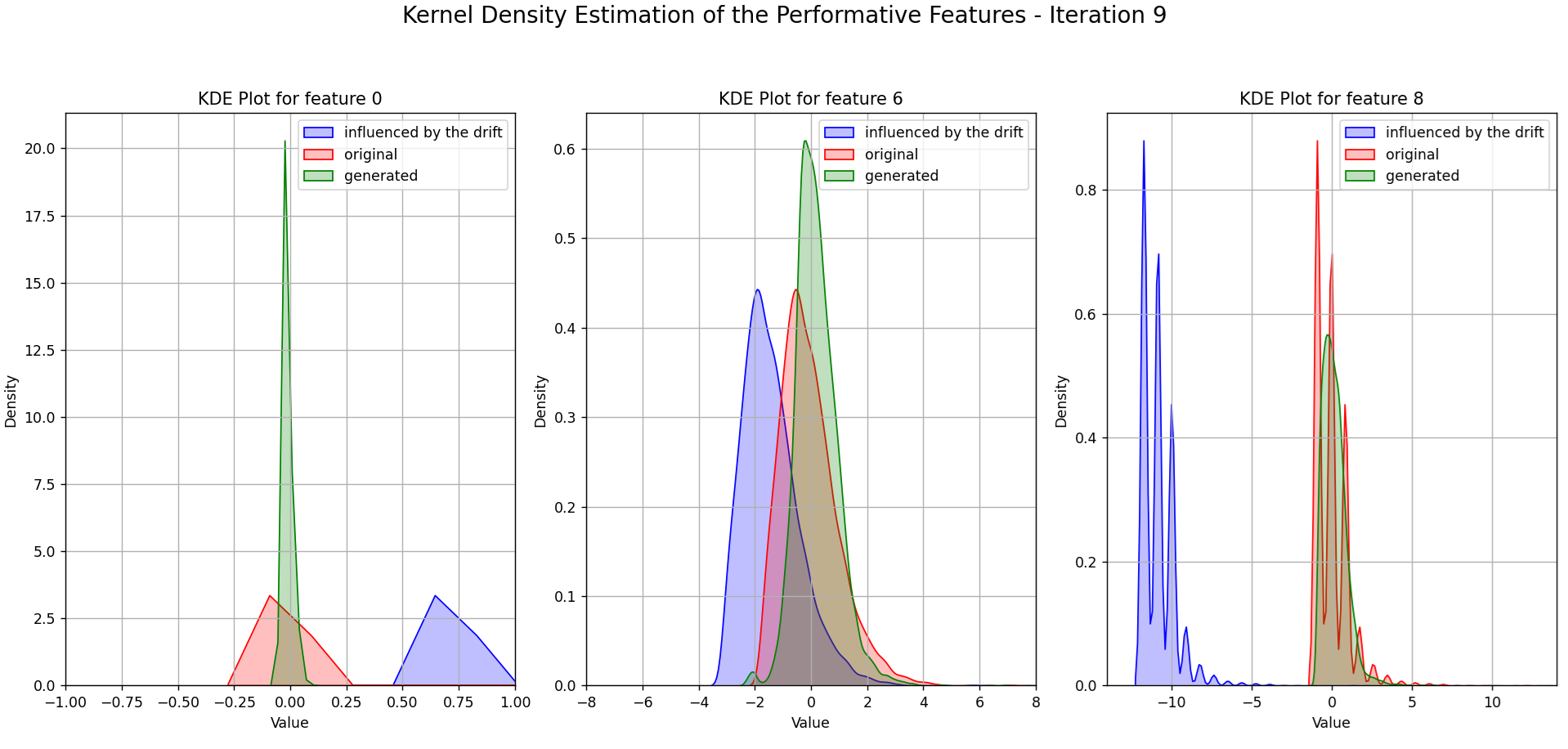}
}

\caption{Kernel Density Estimation of the Perdomo data generator's performative features. Using GDAN, the blue (drifted) data points are projected back to the red (original) data distribution. These projections are indicated by the green areas which show that GDAN is capable of reversing the effects of the drift, despite not being able to recreate the shape of the original distribution exactly.}
\label{fig:Perdomo_KDE_figs}
\end{figure}

Figures \ref{fig:KDE_iter_1} and \ref{fig:KDE_iter_9} illustrate how GDAN acts on on the performative features of the Perdomo data generator. The generated density region (green) does not overlap perfectly with the original one (red). However, GDAN's generator projects the mean of each distribution correctly. Despite GDAN's struggles to replicate outliers, the centres of the green and red density regions remain relatively close, suggesting alignment of central tendencies. A slight degradation occurs over the course of a simulation run, which is visible when analysing feature 8 where two spikes are visible on the green curve at iteration 0, while this same area is more uniform at iteration 9. This explains the decrease in performance observed for both the $M_{LC}$ and $M_G$ (Table \ref{tab:results}) models.

\begin{figure}[t]

\centering
\subfloat[Iteration 1]{\label{fig:Izzo_iter_1}
\centering
\includegraphics[width=0.24\linewidth]{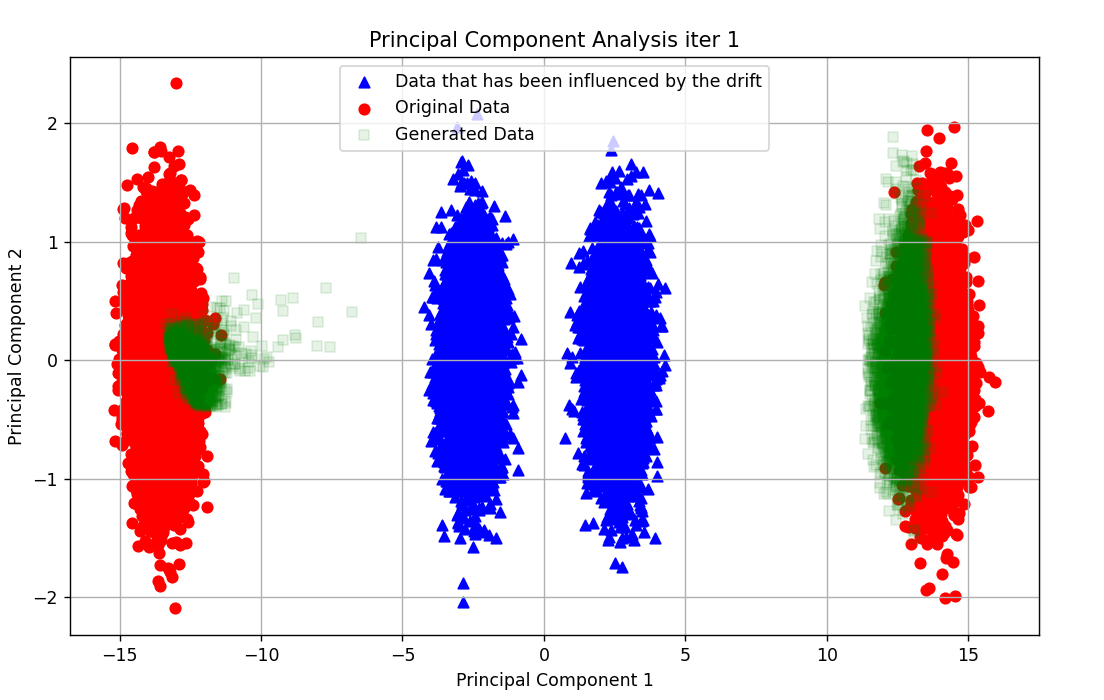}
}
\subfloat[Iteration 4]{\label{fig:Izzo_iter_4}
\centering

\includegraphics[width=0.24\linewidth]{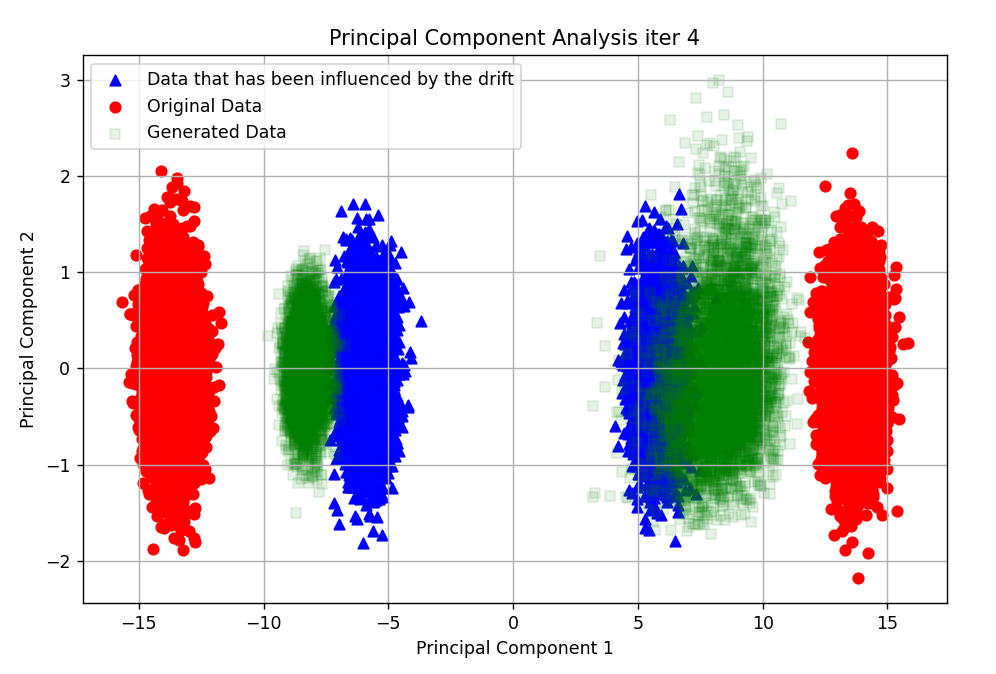}
}
\centering
\subfloat[Iteration 7]{\label{fig:Izzo_iter_7}
\centering
\includegraphics[width=0.24\linewidth]{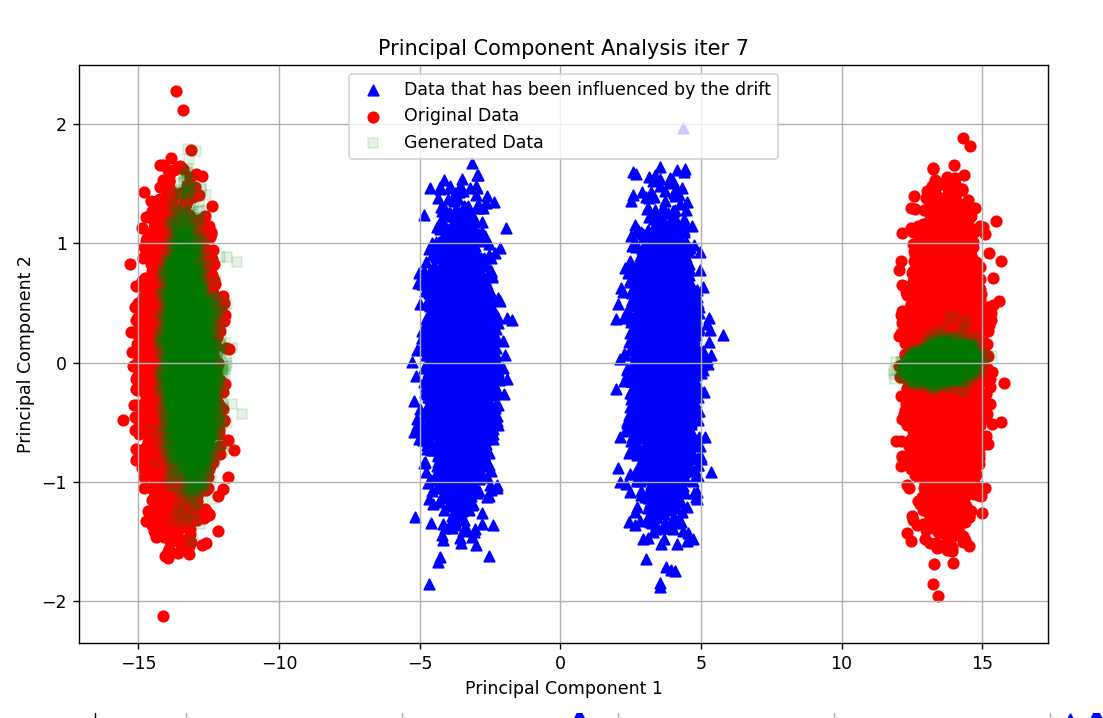}
}
\subfloat[Iteration 10]{\label{fig:Izzo_iter_10}
\centering

\includegraphics[width=0.24\linewidth]{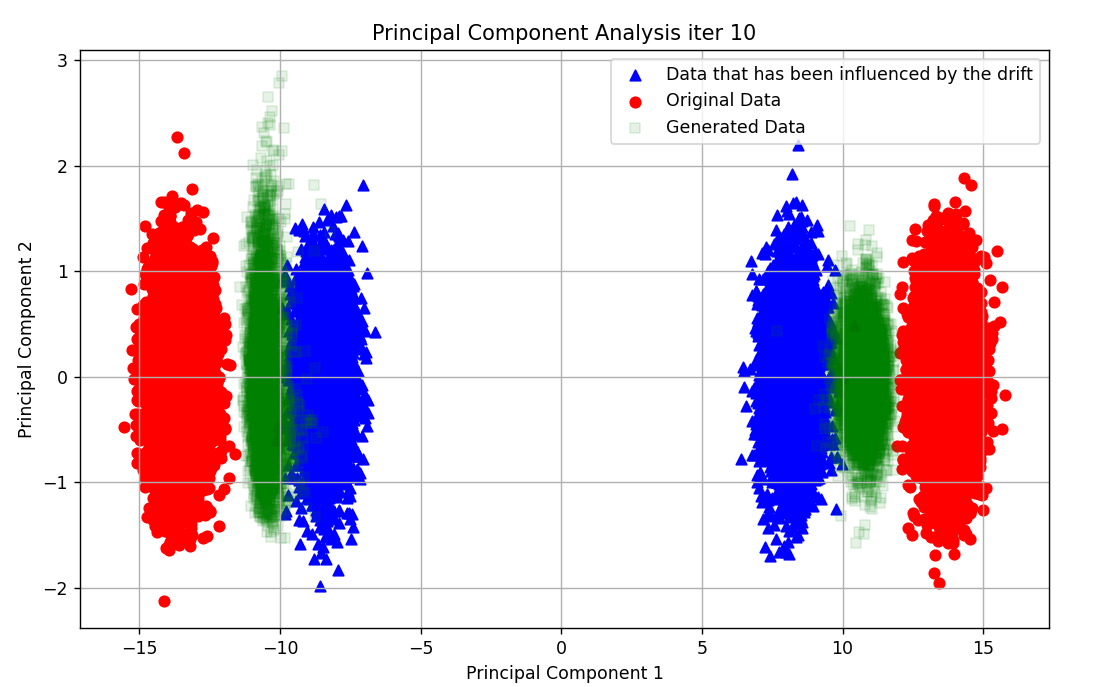}
}

\caption{PCA visualization of the Izzo data generator on iterations where the logistic regressor was retrained. The figure illustrates the cyclical nature of the drift, as the blue (drifted) clusters oscillate, moving to and from the stationary red (original data distribution) clusters. This dynamic significantly affects GDAN's ability to project the drifted data points as shown by the green clusters.}
\label{fig:Izzo_pca}
\end{figure}

Figure \ref{fig:Izzo_pca} presents the evolution of the PCA clusters in the Izzo experiment where the blue (drifted) clusters oscillate between the red (original) ones. Indicated by the inconsistent resemblance of the green (generated) clusters to the original (red) clusters, this oscillatory phenomena of the Izzo data generator impacts the results of GDAN's generator. This is because GDAN is only trained over a single iteration, making it incapable of understanding, and reversing the effects of multi-iteration drift phenomena.

\subsection{Sensitivity of Dataset Size}

\begin{table}[t] 
\centering
\caption{Results of varying dataset sizes with the Perdomo data generator. In general, GDAN performs is incapable of creating a domain-invariant (drift-resistant) representation of the data with fewer training samples.}
\resizebox{\textwidth}{!}{
\begin{tabular}{@{}p{1.8cm}|p{3cm}|p{1cm}p{1cm}p{1cm}p{1cm}p{1cm}p{1cm}p{1cm}p{1cm}p{1cm}p{1cm}@{}}
\toprule
\textbf{No. data points} & \textbf{Metric}
& \textbf{1} & \textbf{2} & \textbf{3} & \textbf{4} & \textbf{5} & \textbf{6} & \textbf{7} & \textbf{8} & \textbf{9} & \textbf{10} \\ \midrule

\multirow{2}{*}{\parbox{2.5cm}{\centering \textbf{5k}}}

    & Acc $M_g$ &
        \makecell{\textbf{71.65} \\ $\pm0.35$} & \makecell{\textbf{71.67} \\ $\pm0.45$} & \makecell{\textbf{71.45} \\ $\pm0.44$} & \makecell{\textbf{70.98} \\ $\pm1.20$} & \makecell{\textbf{70.08} \\ $\pm1.77$} & \makecell{\textbf{68.58} \\ $\pm1.97$} & \makecell{\textbf{66.62} \\ $\pm2.00$} & \makecell{\textbf{65.05} \\ $\pm1.71$} & \makecell{\textbf{63.62} \\ $\pm1.93$} & \makecell{\textbf{62.18} \\ $\pm3.12$} \\
        
    & Acc $M_{LC}$ & 
        \makecell{\textbf{61.16} \\ $\pm0.44$} & \makecell{\textbf{60.74} \\ $\pm0.47$} & \makecell{\textbf{60.47} \\ $\pm0.55$} & \makecell{\textbf{60.36} \\ $\pm0.65$} & \makecell{\textbf{60.54} \\ $\pm0.97$} & \makecell{\textbf{60.84} \\ $\pm1.33$} & \makecell{\textbf{60.80} \\ $\pm1.50$} & \makecell{\textbf{60.74} \\ $\pm1.58$} & \makecell{\textbf{60.69} \\ $\pm1.49$} & \makecell{\textbf{60.34} \\ $\pm1.58$} \\
\midrule
\multirow{2}{*}{\parbox{2.5cm}{\centering \textbf{10k}}}

    & Acc $M_g$ & 
        \makecell{\textbf{71.61} \\ $\pm0.15$} & \makecell{\textbf{71.05} \\ $\pm0.41$} & \makecell{\textbf{69.88} \\ $\pm0.55$} & \makecell{\textbf{68.25} \\ $\pm0.71$} & \makecell{\textbf{66.38} \\ $\pm1.09$} & \makecell{\textbf{64.41} \\ $\pm1.68$} & \makecell{\textbf{62.64} \\ $\pm1.93$} & \makecell{\textbf{61.01} \\ $\pm1.92$} & \makecell{\textbf{59.81} \\ $\pm2.01$} & \makecell{\textbf{58.75} \\ $\pm2.04$} \\
        
    & Acc $M_{LC}$ & 
        \makecell{\textbf{62.47} \\ $\pm0.21$} & \makecell{\textbf{62.82} \\ $\pm0.33$} & \makecell{\textbf{63.41} \\ $\pm0.41$} & \makecell{\textbf{63.74} \\ $\pm0.40$} & \makecell{\textbf{64.16} \\ $\pm0.55$} & \makecell{\textbf{65.02} \\ $\pm0.85$} & \makecell{\textbf{66.27} \\ $\pm1.17$} & \makecell{\textbf{67.52} \\ $\pm1.52$} & \makecell{\textbf{68.50} \\ $\pm1.59$} & \makecell{\textbf{69.26} \\ $\pm1.67$} \\
\bottomrule
\end{tabular}
\label{tab:Perdomo_diff_data}
}
\end{table}

Tables \ref{tab:Perdomo_diff_data} and \ref{tab:Izzo_diff_data} present the results of additional experiments conducted to ascertain GDAN's sensitivity to the size of the training set. The experiment using the Perdomo generator demonstrated that a reduction in training data size negatively affects accuracy metrics. Models trained on $5k$ and $10k$ samples perform worse and performance decays more rapidly. With fewer data points, GDAN struggles to create a domain-invariant representation, which is visible when comparing the accuracies of $M_{LC}$. A spike in the performance of $M_{LC}$ is observed when GDAN is trained on $10k$ samples. This spike is temporary and arises because the model fails to learn the correct direction of drift for one of the performative features (feature 6).
The results of the Izzo generator are more straightforward. The original results (Table \ref{tab:Izzo_results}) were obtained using a model trained on $20k$ samples. The results in Table \ref{tab:Izzo_diff_data} are inferior to those. The model trained on $5k$ data points achieves an accuracy marginally above random guessing, indicating severe underfitting. The model trained on $40k$ data points exhibits clear signs of overfitting, failing to generalize effectively on drifted distributions.
Overall, these results show that GDAN, like most ML-models, is susceptible to performance degradation as the number of training samples tends to zero.

\begin{table}[t]
    \caption{Results of varying dataset sizes with the Izzo data generator. In both cases, GDAN's performance is inferior exhibiting both over (40k) and underfitting (5k). $M_{LC}$ and $M_{ret}$ are omitted as they achieve perfect scores.}
    \centering
    \resizebox{\textwidth}{!}{
    \begin{tabular}{@{}p{2cm}|p{2.5cm}|p{1cm}p{1cm}p{1cm}p{1cm}p{1cm}p{1cm}p{1cm}p{1cm}p{1cm}p{1cm}@{}}
    \toprule
        \centering No. data points&\centering Metric/iter & \centering1&\centering2&\centering3&\centering4&\centering5&\centering6&\centering7&\centering8&\centering9&\makecell{\centering 10} \\
        \midrule
        \multirow{4}{*}{\parbox{2.5cm}{\centering \textbf{5k}}} &\makecell{\raggedright Acc $M_0$} &     
            \makecell{\textbf{50.01} \\ $\pm0.490$} & \makecell{\textbf{49.91} \\ $\pm0.350$} & \makecell{\textbf{49.76} \\ $\pm0.570$} & \makecell{\textbf{50.03} \\ $\pm0.570$} & \makecell{\textbf{54.92} \\ $\pm15.03$} & \makecell{\textbf{54.96} \\ $\pm15.02$} & \makecell{\textbf{54.81} \\ $\pm15.07$} & \makecell{\textbf{49.94} \\ $\pm0.400$} & \makecell{\textbf{50.14} \\ $\pm0.540$} & \makecell{\textbf{49.94} \\ $\pm0.420$} \\
            
        & \makecell{\raggedright Acc $M_g$} &
            \makecell{\textbf{50.01} \\ $\pm0.490$} & \makecell{\textbf{49.91} \\ $\pm0.350$} & \makecell{\textbf{49.76} \\ $\pm0.570$} & \makecell{\textbf{50.03} \\ $\pm0.570$} & \makecell{\textbf{57.84} \\ $\pm16.73$} & \makecell{\textbf{57.85} \\ $\pm16.75$} & \makecell{\textbf{57.75} \\ $\pm16.77$} & \makecell{\textbf{49.95} \\ $\pm0.400$} & \makecell{\textbf{50.14} \\ $\pm0.540$} & \makecell{\textbf{49.94} \\ $\pm0.420$} \\
        \midrule
         \multirow{4}{*}{\parbox{2.5cm}{\centering \textbf{40k}}} & \makecell{\raggedright Acc $M_0$} &
            \makecell{\textbf{49.95} \\ $\pm0.580$} & \makecell{\textbf{49.99} \\ $\pm0.520$} & \makecell{\textbf{49.91} \\ $\pm0.510$} & \makecell{\textbf{49.84} \\ $\pm0.470$} & \makecell{\textbf{55.86} \\ $\pm15.32$} & \makecell{\textbf{55.79} \\ $\pm15.32$} & \makecell{\textbf{56.10} \\ $\pm15.21$} & \makecell{\textbf{50.00} \\ $\pm0.480$} & \makecell{\textbf{49.98} \\ $\pm0.460$} & \makecell{\textbf{50.00} \\ $\pm0.390$} \\
            
        & \makecell{\raggedright Acc $M_g$} &
            \makecell{\textbf{68.10} \\ $\pm10.95$} & \makecell{\textbf{68.25} \\ $\pm10.92$} & \makecell{\textbf{68.09} \\ $\pm10.85$} & \makecell{\textbf{68.16} \\ $\pm10.96$} & \makecell{\textbf{58.76} \\ $\pm18.26$} & \makecell{\textbf{58.75} \\ $\pm18.28$} & \makecell{\textbf{59.02} \\ $\pm18.14$} & \makecell{\textbf{68.63} \\ $\pm10.89$} & \makecell{\textbf{68.75} \\ $\pm10.90$} & \makecell{\textbf{68.80} \\ $\pm10.89$} \\
        \bottomrule
    \end{tabular}
    }
    \label{tab:Izzo_diff_data}
\end{table}

\section{Conclusions and Future Work}

In this work we addressed the problem of performance degradation in settings with performative concept drift. That is, scenarios where a model's predictions influence the data it must later predict on. We developed GDAN, a novel architecture that combines Generative and Domain Adversarial Neural Networks. By creating domain-invariant representations of incoming data, GDAN is able to reverse the effects of performative drift and enable stable classification. 
Empirically evaluating GDAN on the Perdomo \cite{PerdomoZrnicMendlerdunnerEtal2020} and Izzo \cite{IzzoZacharyZou2022} data generators, results indicated that GDAN performs sufficiently well, consistently outperforming baseline models trained under the same conditions. Our results also indicated GDAN's performance degrades in scenarios where the direction of performative drift changes frequently. In such scenarios, GDAN can be inferior to simple retraining regimes. However, there are cases where retraining is not feasible, such as when new labels arrive with delay. We also showed that GDAN can model the direction of performative drift, and reverse its effects. This allows GDAN to provide deeper insights into the nature of performative drift, and allows other models to leverage GDAN to make themselves drift-resistant.
Future work will aim to use real-world data to validate GDAN's performance as the data generators used in this work only approximate real-world phenomena. Furthermore, this work only investigated scenarios with performative drift whereas real world settings may have both performative and traditional (intrinsic) concept drift. Extending GDAN to include performative drift detection methods \cite{GowerWinterKrempl2025} would also be beneficial. This would address one the strongest assumptions in Performative Prediction literature which is that the presence of performative drift is known beforehand. Lastly, performative drift does not behave like intrinsic drift. Unlike intrinsic drift, the volatility of performative drift is tied to the rate at which deployed predictors are updated. Our results show that in settings that are performative, incremental learning strategies may be less effective as they result in more volatile drift phenomena. Future work will examine these claims in greater detail and seek to propose alternative learning strategies.

\bibliographystyle{unsrt}  
\bibliography{main}  

\appendix

\begin{table}[!h]
\centering
\caption{Baseline experiment whereby data is influenced once and then continuously sampled from distribution $t_1$. In this scenario, no deterioration occurs and the performance of $M_{LC}$ and $M_g$ are superior to the accuracies achieved by $M_0$ and $M_{ret}$. This experiment serves as a sanity check for our proposed solution.}
\resizebox{\textwidth}{!}{
\begin{tabular}{@{}p{3cm}|p{1cm}p{1cm}p{1cm}p{1cm}p{1cm}p{1cm}p{1cm}p{1cm}p{1cm}p{1cm}@{}}
\toprule
\textbf{Metric}
& \textbf{1} & \textbf{2} & \textbf{3} & \textbf{4} & \textbf{5} & \textbf{6} & \textbf{7} & \textbf{8} & \textbf{9} & \textbf{10} \\ \midrule
 Acc $M_0$ & \makecell{\textbf{71.62} \\ $\pm0.23$} & \makecell{\textbf{71.62} \\ $\pm0.23$} & \makecell{\textbf{71.62} \\ $\pm0.23$} & \makecell{\textbf{71.62} \\ $\pm0.23$} & \makecell{\textbf{71.62} \\ $\pm0.23$} & \makecell{\textbf{71.62} \\ $\pm0.23$} & \makecell{\textbf{71.62} \\ $\pm0.23$} & \makecell{\textbf{71.62} \\ $\pm0.23$} & \makecell{\textbf{71.62} \\ $\pm0.23$} & \makecell{\textbf{71.62} \\ $\pm0.23$} \\
 Acc $M_{ret}$ & \makecell{\textbf{70.69} \\ $\pm0.98$} & \makecell{\textbf{70.69} \\ $\pm0.98$} & \makecell{\textbf{70.69} \\ $\pm0.98$} & \makecell{\textbf{70.69} \\ $\pm0.98$} & \makecell{\textbf{70.69} \\ $\pm0.98$} & \makecell{\textbf{70.69} \\ $\pm0.98$} & \makecell{\textbf{70.69} \\ $\pm0.98$} & \makecell{\textbf{70.69} \\ $\pm0.98$} & \makecell{\textbf{70.69} \\ $\pm0.98$} & \makecell{\textbf{70.69} \\ $\pm0.98$} \\
 Acc $M_g$ & \makecell{\textbf{72.76} \\ $\pm0.20$} & \makecell{\textbf{72.76} \\ $\pm0.20$} & \makecell{\textbf{72.76} \\ $\pm0.20$} & \makecell{\textbf{72.76} \\ $\pm0.20$} & \makecell{\textbf{72.76} \\ $\pm0.20$} & \makecell{\textbf{72.76} \\ $\pm0.20$} & \makecell{\textbf{72.76} \\ $\pm0.20$} & \makecell{\textbf{72.76} \\ $\pm0.20$} & \makecell{\textbf{72.76} \\ $\pm0.20$} & \makecell{\textbf{72.76} \\ $\pm0.20$} \\
Acc $M_{LC}$ & \makecell{\textbf{76.00} \\ $\pm0.20$} & \makecell{\textbf{76.00} \\ $\pm0.20$} & \makecell{\textbf{76.00} \\ $\pm0.20$} & \makecell{\textbf{76.00} \\ $\pm0.20$} & \makecell{\textbf{76.00} \\ $\pm0.20$} & \makecell{\textbf{76.00} \\ $\pm0.20$} & \makecell{\textbf{76.00} \\ $\pm0.20$} & \makecell{\textbf{76.00} \\ $\pm0.20$} & \makecell{\textbf{76.00} \\ $\pm0.20$} & \makecell{\textbf{76.00} \\ $\pm0.20$} \\
$L_1(X_0, X_i)$ & \makecell{\textbf{0.107} \\ $\pm0.000$} & \makecell{\textbf{0.107} \\ $\pm0.000$} & \makecell{\textbf{0.107} \\ $\pm0.000$} & \makecell{\textbf{0.107} \\ $\pm0.000$} & \makecell{\textbf{0.107} \\ $\pm0.000$} & \makecell{\textbf{0.107} \\ $\pm0.000$} & \makecell{\textbf{0.107} \\ $\pm0.000$} & \makecell{\textbf{0.107} \\ $\pm0.000$} & \makecell{\textbf{0.107} \\ $\pm0.000$} & \makecell{\textbf{0.107} \\ $\pm0.000$} \\
$L_1(X_0,G(F(X_i))$ & \makecell{\textbf{0.112} \\ $\pm0.001$} & \makecell{\textbf{0.112} \\ $\pm0.001$} & \makecell{\textbf{0.112} \\ $\pm0.001$} & \makecell{\textbf{0.112} \\ $\pm0.001$} & \makecell{\textbf{0.112} \\ $\pm0.001$} & \makecell{\textbf{0.112} \\ $\pm0.001$} & \makecell{\textbf{0.112} \\ $\pm0.001$} & \makecell{\textbf{0.112} \\ $\pm0.001$} & \makecell{\textbf{0.112} \\ $\pm0.001$} & \makecell{\textbf{0.112} \\ $\pm0.001$} \\

\bottomrule
\end{tabular}}
\label{tab:BENCHMARK}
\end{table}

\section{Training GDAN} \label{sect:PSEUDOCODE}
Algorithm \ref{alg:training_GDANN} highlights the process of training GDAN. $\theta$ denotes the weights of a model, with the subscript indicating the specific model. The parameter $\mu$ is the model specific learning rate. $\lambda_1$ and $\lambda_2$ are regularisation parameters, which control the degree of penalisation a model receives. The exact values for training GDAN in this work are included in the Supplementary Materials.

\newcommand{\dotstep}{\vspace{0.5em} \item[\hspace{2.5em}$\bullet$]}
\newcommand{\ddotstep}{\vspace{0.5em} \item[\hspace{4em}$\bullet$]}

\begin{algorithm}[!h]
\caption{Gradient Descent Training of GDAN}
\begin{algorithmic}
\State \textbf{Input:} Training iterations $N$, batch size $m$ and $k$ (the number of steps after which the generator starts to be updated and the feature extractor stops being updated)\\

\State Create $Z = X_0 \cup X_1$ with class labels $C$ and domain labels $D_l$ \\

\For{number of training iterations}
    \dotstep Sample minibatch of $\frac{m}{2}m$ examples $\{x^{(1)}, \ldots, x^{(\frac{m}{2})}\}$ from $Z$
    \dotstep Arbitrarily pair each sample with a point from distribution $X_0$
    \dotstep Pass each sample through the network components and record their outputs.
    \dotstep Generate another minibatch of $\frac{m}{2}$ samples by using the generator with $F(x^{(\frac{m}{2})})$ as input, and assign domain labels. The generator loss will only be calculated based on these generated instances (later denoted as $m_1$). The discriminator is trained on both real and generated samples.
    \dotstep Calculate loss functions.
    \ddotstep Update $F$, $LC$, $D$, and $G$:
    \[
    \theta_F \leftarrow \theta_F - \mu_F (\nabla_{\theta_F} \mathscr{L}_{LC}^m - \lambda_1 \nabla_{\theta_F} \mathscr{L}_{domain}^{\frac{1}{2}m})
    \]
    \[
    \theta_{LC} \leftarrow \theta_{LC} - \mu_{LC} \nabla_{\theta_{LC}} \mathscr{L}_{LC}^m
    \]
    \[
    \theta_D \leftarrow \theta_D - \mu_D \nabla_{\theta_D} (\mathscr{L}_{source}^m + \mathscr{L}_{domain}^m)
    \]
    \[
    \theta_G \leftarrow \theta_G - \mu_G \nabla_{\theta_G} (\mathscr{L}_{domain}^{m_1} + \mathscr{L}_{source}^{m_1} + \lambda_2 \mathscr{L}_{L1})
    \]
\EndFor \\
\State \textbf{Output:} Trained $F$, $LC$, $G$, and $D$
\end{algorithmic}
\label{alg:training_GDANN}
\end{algorithm}

\end{document}